%% file: 0953.tex
\documentclass{bmvc2k}

\usepackage{graphicx}
\usepackage{amsmath}
\usepackage{amssymb}

\usepackage[utf8]{inputenc} 
\usepackage[T1]{fontenc}    
\usepackage{url}            
\usepackage{amsfonts}       
\usepackage{graphicx}
\usepackage{graphics}
\usepackage{bm}
\usepackage{multirow}

\usepackage{multicol}

\title{Learning Short-Cut Connections for Object Counting}

\addauthor{Daniel O\~noro-Rubio}{daniel.onoro@neclab.eu}{1}
\addauthor{Mathias Niepert}{mathias.niepert@neclab.eu}{1}
\addauthor{Roberto J. L\'opez-Sastre}{robertoj.lopez@uah.es}{2}

\addinstitution{
 SysML,\\
 NEC Lab Europe,\\
 Heidelberg, Germany
}
\addinstitution{
 GRAM,\\
 University of Alcal\'a,\\
 Alcal\'a de Henares, Spain
}

\runninghead{O\~noro-Rubio, Niepert, L\'opez-Sastre}{Learning Short-Cut Connections\ldots}

\def\etal{\emph{et al}\bmvaOneDot}
\def\ie{\emph{i.e}\bmvaOneDot}

\usepackage[colorlinks]{hyperref}

\begin{document}

\maketitle

\input{abstract}

\input{introduction}

\input{related}

\input{model}

\input{architecture}

\input{experiments}

\input{conclusion}

\bibliography{crowd_ref}
\end{document}

%% file: abstract.tex
\begin{abstract}
Object counting is an important task in computer vision due to its growing demand in applications such as traffic monitoring or surveillance. In this paper, we consider object counting as a learning problem of a joint feature extraction and pixel-wise object density estimation with Convolutional-Deconvolutional networks. We introduce a novel counting model, named Gated U-Net (GU-Net). Specifically, we propose to enrich the U-Net architecture with the concept of \emph{learnable} short-cut connections. Standard short-cut connections are connections between layers in deep neural networks which skip at least one intermediate layer. Instead of simply setting short-cut connections, we propose to learn these connections from data. Therefore, our short-cuts can work as gating units, which optimize the flow of information between convolutional and deconvolutional layers in the U-Net architecture. We evaluate the introduced GU-Net architecture on three commonly used benchmark data sets for object counting. GU-Nets consistently outperform the base U-Net architecture, and achieve state-of-the-art performance.
\end{abstract}

%% file: introduction.tex
\section{Introduction}

Counting objects in images is an important problem with numerous applications. Counting people in crowds~\cite{idrees2013}, cells in a microscopy image~\cite{cellsDataset},  or vehicles in a traffic jam \cite{guerrero2015} are typical instances of this problem. Computer vision systems that automate these tasks have the potential to reduce costs and processing time of otherwise labor intensive tasks.

An intuitive approach to object counting is through object detection~\cite{dalal2005,Wu2005,felzenszwalb2010}. The accuracy of object localization approaches, however, rapidly deteriorates as the density of the objects increases. Thus, methods that approach the counting problem as one of object density estimation, following the pioneering work of Lempitsky \etal \cite{lempitsky2010}, have been shown to achieve state-of-the-art results~\cite{arteta2014,zhang_2015_CVPR,Onoro_2016_ECCV,zhang2016}.

\begin{figure}
\centering
\includegraphics[width=1.0\textwidth]{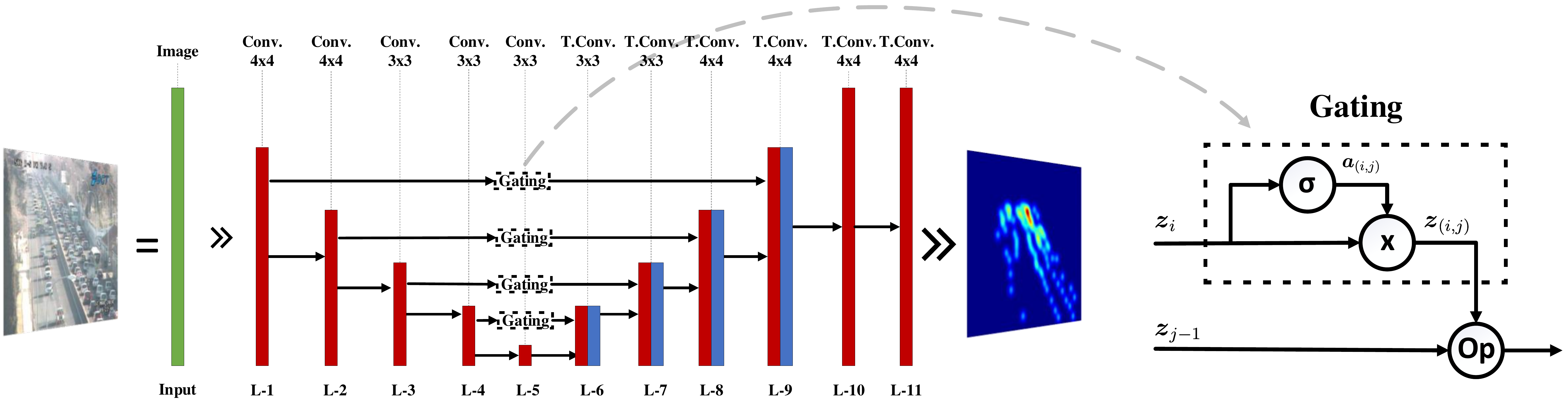}
\caption{\label{fig:gated-architecture} Proposed Gated U-Net Network architecture, where each gating unit is a \emph{learnable} short-cut connection between convolutional and deconvolutional layers.}
\end{figure}

With this paper, we introduce a deep learning model that generates object density maps for counting problems. Figure~\ref{fig:gated-architecture} illustrates the architecture which we refer to as the Gated U-Net (GU-Net). We propose the use of \emph{learnable} short-cut connections in the context of neural architectures with deconvolutional layers, such as the U-Net \cite{unetRonnebergerFB15} and V-net \cite{MilletariNA16}. Short-cut connections are connections between layers that skip at least one intermediate layer. There is an increasing body of work on short-cut connections (sometimes also referred to as skip-connections) in deep neural networks. Examples of architectures with short-cut connections are: highway~\cite{Srivastava_2015_NIPS} and residual networks~\cite{he2016deep}, DenseNets~\cite{Huang_2017_CVPR}, and U-Nets~\cite{unetRonnebergerFB15}. While a clear benefit of these connections has been shown repeatedly, the exact reasons for their success are still not fully understood. Short-cut connections allow one to train very deep networks by facilitating a more direct flow of information between far apart layers during backpropagation~\cite{he2016deep}. A more recent and complementary explanation is that networks with short-cut connections behave similar to ensembles of shallower networks~\cite{veit16residual}.


Instead of simply setting short-cut connections, in this paper, we propose to \emph{learn} these connections from data. Therefore, our learnable short-cut connection can work as a gating unit, which optimizes the flow of information between convolutional and deconvolutional layers (see again Figure \ref{fig:gated-architecture}).

The main contributions of our work are as follows:
\begin{itemize}
 \item We first introduce, in Section \ref{sec:gating_units}, our learnable short-cut connections, \ie the Gating units, as a general bypass mechanism applicable to any neural network. 
 \item In Section \ref{sec:counting_model}, we explore the U-Net architecture~\cite{unetRonnebergerFB15}, a popular model for image segmentation problems, as a reference approach for object counting. We extend the U-Net model by integrating the proposed Gating units. The resulting GU-Net architecture is able to learn which feature maps are beneficial for deconvolutional layers, and also the extent to which feature maps should be passed to later layers.
 \item We conduct an extensive set of experiments (Section \ref{sec:experiments}) that explore the benefits of the proposed model for the problems of crowd and vehicles counting.
\end{itemize}

%% file: related.tex
\section{Related Work}

In recent years, the benefits of short-cut connections in deep neural networks have been demonstrated in various empirical studies~\cite{Bell_2016_CVPR, Hariharan_2015_CVPR, Sermanet_2013_ICCV_Workshops, Yang_2015_ICCV, he2016deep, Huang_2017_CVPR}. Initial research focuses on augmenting top layers with feature maps from bottom layers, where the RCNN framework~\cite{girshick2014} is modified by connecting the last convolutional layers of the VGG16 architecture using a concatenation operation~\cite{Bell_2016_CVPR}. In later works, the importance of short-cut connections for back-propagating the gradients during training, and its mitigating effect on the vanishing gradient problem, have been emphasized. \cite{Srivastava_2015_NIPS} propose ``high way'' networks which, inspired by LSTMs~\cite{lstm1999}, have an architecture where the input of a layer is the sum of the gated input of the previous layer and the gated output of the previous layer. Following a similar strategy, residual networks~\cite{he2016deep} and DenseNets~\cite{Huang_2017_CVPR} take advantage of multiple shortcut connections. 

More recent work introduces a gating mechanism that operates directly on the short-cut connections~\cite{Islam_2017_CVPR}. The proposed gated refinement network predicts the ground truth at multiple scales, where a superior scale is predicted based on the concatenation of the prediction of the previous scale, and the output of its corresponding gating unit. On their gating units, the feature maps of two consecutive scales are combined creating a totally new feature map that is used to refine the prediction. 

In contrast to previous work, we present a novel self-gating mechanism that performs a spatial attention over the information forwarded through the short-cut connections. The presented method improves the results on models with a single output, therefore further refinement is still allowed afterward.

We refer the reader to the survey of prior work on object counting of Loy \etal \cite{loy2013}. In summary, different type of counting techniques have been applied such as counting by detection \cite{dalal2005,felzenszwalb2010,chen2015,leibe2005,li2008,patzold2010,viola2004,wang2011}, counting by clustering \cite{babaud2006,tu2008}, and counting by regression \cite{arteta2014,chan2008,fiaschi2012,lempitsky2010,zhang_2015_CVPR,rodriguez2011,pham2015}. At this point in time, the best-performing methods belong to the group of counting by regression which was introduced by Lempitsky \etal \cite{lempitsky2010}. There a method for learning a linear mapping from local image features to object density maps is introduced. Fiaschi et al. \cite{fiaschi2012} use a random forest trained with the aid of a collection of multi-spectral filters. In \cite{zhang_2015_CVPR} they propose 5 layers convolutional neural network trained with a switching loss function. The work \cite{zhang2016} proposes a neural that combines in parallel multiple filter sizes. In \cite{Sam_2017_CVPR} they present a framework where a neural network analyzes the image by region and selects an existing model to estimate the output. In the work of \cite{Zhang_2017_CVPR} they divide the image, and they train a deep neural network that contains a regressor for each region.  In \cite{Sindagi_2017_ICCV} they present a system that combines local and global features by aggregating multiple neural networks trained with a reconstruction loss and an adversarial loss. In Yuhong \etal~\cite{yuhong2018_csrnet} they extend the VGG16 with the dilated convolution \citep{YuKoltun2016} achieving the top performance of the state of the art. With the presented work we proposed a general gating mechanism that can be easily implemented on different models. In our experiments we show how it improves the performance of our U-Net baseline model. Moreover, the proposed models achieve the state of the art performance without requiring a pretraining on large datasets such as ImageNet and with just a fraction of the parameters of the top performing methods.

%% file: model.tex
\section{Learnable Short-Cut Connections}
\label{sec:gating_units}

Short-cut connections are links between layers in deep neural networks (DNN) which skip at least one intermediate layer. We here introduce a framework for fine-grained learning of the strength of those connections. More specifically, we propose a gating mechanism that can be applied to any deep neural network, to learn short-cuts from the training data, hence optimizing the flow of information between earlier and later layers of the architecture. 

We define a DNN as a function approximator that maps feature tensors to output tensors,
\begin{equation}
\label{eq:dnn}
\boldsymbol{y} = F_W(\boldsymbol{x}),
\end{equation} 
where $W$ represents the parameters of the model. 

A DNN $F_{W}(\boldsymbol{x})$ can be written as a composition of functions $F_W(\boldsymbol{x}) = f_{W_n}^n \circ \dots f_{W_2}^2 \circ f_{W_1}^1(\boldsymbol{x})$, where each $f^i_{W_i}$ represents the transformation function for layer $i \in \{1, ..., n\}$ with its parameters $W_i$. Each such layer maps its input to an output tensor $\boldsymbol{z}_{i}$. Typically, the output $\boldsymbol{z}_{i}$ of layer $i$  is used as input to the proceeding layer $f^{i+1}_{W_{i+1}}(\boldsymbol{z}_{i})$ to generate output $\boldsymbol{z}_{i+1}$, and so on. Each layer, therefore, is only connected locally to its preceding and proceeding layers.

However, we can also connect the output of a layer $i$ to the input of a layer $j$, being $j > i + 1$, that is, we can create \emph{short-cut connections} that skip a number of layers in the network. The first class of network architectures popularizing these skip-connections have been introduced with the ResNet model class~\cite{he2016deep}. Here, we do not only create certain short-cut connections, but \emph{learn} their connection strength using a \emph{gated short-cut unit}, see Figure \ref{fig:gated-architecture}. The short-cut units determine the amount of information which is passed to other layers, and also the ways in which this information is combined with the input of these later layers. 

Whenever a layer $i$ with transformation function $f^{i}_{W_i}$ is connected to a layer $j$, with $j > i + 1$, we introduce a new convolutional layer $g^{(i,j)}$, whose hyper parameters are identical to that of layer $i$. With $\boldsymbol{z}_i$ being the output of layer $i$, we compute 
\begin{equation}
\label{eq:gated_op}
\boldsymbol{a}_{(i,j)} = \sigma(g^{(i,j)}(\boldsymbol{z}_{i})),
\end{equation} 
where $\sigma$ is the sigmoid function. Tensor $\boldsymbol{a}_{(i,j)}$ consists of scalars between $0$ and $1$ that determine the amount of information passed through to layer $j$, for each of the local convolutions applied to $\boldsymbol{z}_{i}$. We then compute the element-wise product of the output of layer $i$ and the tensor $\boldsymbol{a}_{(i,j)}$, as follows,
\begin{equation}
\label{eq:gated_feat}
\boldsymbol{z}_{(i, j)} = \boldsymbol{z}_i \odot \boldsymbol{a}_{(i,j)},
\end{equation}
where $\boldsymbol{a}_{(i,j)}$ are again the values of the gating function in charge of deciding how much information is let through.

Figure \ref{fig:gated-architecture} illustrates the gating unit. $\boldsymbol{z}_i$ is used as input to a convolutional layer to compute the gating scores $\boldsymbol{a}_{(i,j)}$. A point-wise multiplication is performed to obtain $\boldsymbol{z}_{(i,j)}$. These gated features are then combined with the input $\boldsymbol{z}_{j-1}$ of layer $j$ by using an operation $\circ$ (sum, point-wise multiplication, or concatenation) between $\boldsymbol{z}_{(i,j)}$ and $\boldsymbol{z}_{j-1}$. The resulting tensor $\boldsymbol{z}_{(i,j)} \circ \boldsymbol{z}_{j-1}$ is then used as the input for layer $j$.

%% file: architecture.tex
\section{GU-Net: Gated Short-Cuts for Object Counting}
\label{sec:counting_model}


A class of network architecture with standard short-cut connections is the U-Net~\cite{unetRonnebergerFB15}. U-Nets are fully convolutional neural networks which can be divided into a compression (convolution) and a construction (deconvolution) part. Both parts are connected with short-cut connections leading from a compression layer to a construction layer with the same size. We chose U-Nets since they have shown good performance on several data sets for image segmentation and since they contain short-cut connections which we aim to learn. 

Figure \ref{fig:gated-architecture} illustrates the DNN architecture we propose for the object counting problem. With the exception of the gating units, which we introduce with this paper, we have a standard U-Net architecture \cite{unetRonnebergerFB15}. The U-Net model consists of 11 layers. The first five layers are convolutional layers that comprise the encoding of the input images. The following five layers are transpose-convolutional layers \cite{Long_2015_CVPR} that perform the reconstruction (or decoding) of the bottleneck representation. In all convolutional and deconvolutional layers we use a stride of $[2,2]$ and we make use of the Leaky ReLu~\cite{andrew2013} as activation function. Finally, the eleventh layer is a convolution operation with a single filter that produces the output.


For our experiments, we have setup up the receptive field or our models to cover an area of $[96,96]$ pixels. Since we test our algorithm for the counting problem, we find out that an area of $[96,96]$ pixels can hold a big object surrounded by background. The size of the filter is chosen to be $[4,4]$ for those features maps that are greater or equal to $[24,24]$, while for those that are smaller we use a slightly reduced size of $[3,3]$. 

In this work, we tackle the problem of object counting with a deep network that is trained to learn how to map the appearance of the images to their corresponding object density maps. Therefore, the goal of our network is to generate image-wise density maps, that can be later integrated to get the total object count which is presented in the image. Therefore, we propose to solve a multivariate regression task, where the network is trained to minimize the squared difference between the prediction and the ground truth. Therefore, we define our loss function as: $\mathcal{L}(\boldsymbol{y}, \boldsymbol{x}) = \left \| \boldsymbol{y} - F_W(\boldsymbol{x}) \right \|_2$. In contrast to most previous works, we can achieve state-of-the-art performance with just a single loss term.

In the GU-Net, the gating units act as bit-wise soft masks that determine the amount of information to pass to the respective layers, so as to improve the quality of the final feature maps. In a deep neural network, the first layers are specialized in detecting low-level features such as edges, textures, colors, etc. These low-level features are needed in a normal feed-forward neural network, but when they are combined with deeper features, they might or might not contribute to the overall performance. For this reason, the gating units are specially useful to automate the feature selection mechanism for improving the \emph{short-cuts} connections, while strongly back propagating the updates to the early layers during learning.

The main limitation of the presented idea is that the gating units do not add more freedom to the general network but add additional parameters (one additional convolutional layer per short-cut connection). Therefore, the fitting capability of GU-Net is the same as the original U-Net. However, the gating strategy leads to more robust models that produce better results.

%% file: experiments.tex
\section{Experiments}
\label{sec:experiments}
We have conducted experiments on three publicly available object counting data sets: \textsc{Trancos}~\cite{guerrero2015}, \textsc{ShanghaiTech}~\cite{zhang2016}, and \textsc{UCSD}~\cite{chan2008}. We perform a detailed comparison with state-of-the-art object counting methods. Moreover, we provide empirical insights into the advantages of the proposed learnable short-cut connections.

\subsection{General Setup}
\label{sec:setup}

We use Tensorflow \cite{tensorflow2015} to implement the proposed models. To train our models, we initialize all the weights with samples from a normal distribution with mean zero and standard deviation $0.02$. We make use of the $L2$ regularizer with a scale value of $2.5 \times 10^{-5}$ in all layers. To perform gradient decent we employ Adam \cite{adamKingmaICLR2015} with a learning rate of $10^{-4}$, $\beta_1$ of $0.9$, and $\beta_2$ of $0.999$. We train our solutions for $2 \times 10^5$ iterations, with mini-batches consisting of $128$ image patches, which are extracted from a random image of the training set, such that $50\%$ of the patches contain a centered object and the remaining patches are randomly sampled from the image. We perform data augmentation by flipping images horizontally with a probability of $0.5$.  All the pixel values from all channels are scaled to the range $[0,1]$. The ground truth of each dataset is generated by placing a normal distribution on top of each of the annotated objects in the image. The standard deviation $\sigma$ of the normal distribution varies depending on the data set under consideration. 

To perform the object count we feed entire images to our models. The proposed models are fully convolutional neural networks, therefore they are not constrained to a fixed input size. Finally, note that the presented models have significantly fewer parameters than most of the state-of-the-art methods. To analyze an image of size $[640,480,3]$ with a GU-Net architecture, takes on average only 23ms when executed in a NVIDIA GeForce 1080 Ti.

We use mean absolute error ($MAE$) and mean squared error ($MSE$) for the evaluation of the results on \textsc{ShanghaiTech}~\cite{zhang2016} and \textsc{UCSD}~\cite{chan2008}: \vspace{-0.9cm} \begin{multicols}{2}
  \begin{equation}
    MAE = \frac{1}{N} \sum^{N}_{n=1} \left| \mathtt{c}(F_W(\boldsymbol{x})) - \mathtt{c}(\boldsymbol{y})  \right|\, ,
  \end{equation}
  \begin{equation}
    MSE = \sqrt[]{\frac{1}{N} \sum^{N}_{n=1} {\left( \mathtt{c}(F_W(\boldsymbol{x})) - \mathtt{c}(\boldsymbol{y})  \right)}^2}\, .
  \end{equation}
\end{multicols}

$N$ is the total number of images, $\mathtt{c}(F_W(\boldsymbol{x}))$ represents the object count based on the density map output of the neural network, and $\mathtt{c}(\boldsymbol{y})$ is the count based on the ground truth density map. 

The Grid Average Mean absolute Error ($\mathtt{GAME}$) metric is used to perform the evaluation with \textsc{Trancos}~\cite{guerrero2015}. According to this metric, each image has to be divided into a set of $4^s$ non-overlapping rectangles. When $s=0$ it is the entire image, when $s=1$ it is the four quadrants, and so on. For a specific value of $s$, the $\mathtt{GAME}$(s) is computed as the average of the MAE in each of the corresponding $4^s$ subregions. Let $\mathbf{R}_{s}$ be the set of image regions resulting from dividing the image into $4^s$ non-overlapping rectangles. Then, the $\mathtt{GAME}$(s) is obtained as follows,
\vspace{-0.25cm}
\begin{equation}
\mathtt{GAME}(s) = \frac{1}{N} \sum^{N}_{n=1} \hspace{-1mm} \left( \frac{1}{|\mathbf{R}_s|} \sum_{\boldsymbol{x} \in \mathbf{R}_s} \hspace{-1mm} \left| \mathtt{c}(F_W(\boldsymbol{x})) - \mathtt{c}(\boldsymbol{y})  \right|\right)\, .
\end{equation}

This way, this metric provides an spatial measure of the error for $s > 0$. Note that $\mathtt{GAME}(0)$ is simply the Mean Absolute Error (MAE).

\subsection{Counting vehicles}

\textsc{Trancos} is a publicly available dataset consisting of images depicting traffic jams in various road scenarios, and under multiple lighting conditions and different perspectives. It provides 1.244 images obtained from video cameras where a total of 46.796 vehicles have been annotated. It also includes a region of interest (ROI) per image.

For the experiments using this dataset, we follow the experimental setup of \cite{guerrero2015} to generate the ground truth density maps by setting the standard deviation of the normal distributions to $\sigma=10$. We also perform a random gamma transformation to the images.

\begin{table}[t!]
\small
\centering
\caption{Fusion operation comparison for the GU-Net model in the \textsc{Trancos} dataset.}
\label{tab:trancos_fusion}
\begin{tabular}{|l|c|c|c|c|}
\hline
Model                           & GAME 0 & GAME 1 & GAME 2 & GAME 3 \\
\hline
\hline
Ew. Multiplication              & 6.19   & 7.24   & 8.64   & 10.51   \\
\hline
Summation                       & 4.81   & 6.09   & 7.63   & 9.60   \\
\hline
Concatenation                   & \textbf{4.44}   & \textbf{5.84}   & \textbf{7.34}   & \textbf{9.29}   \\
\hline
\end{tabular}
\end{table}

We first use this dataset to evaluate the performance of the different possible operations (element-wise multiplication or summation, as well as a simple concatenation) for the operator $\circ$ of our learnable short-cut connections. Table~\ref{tab:trancos_fusion} shows the results for the GU-Net architecture and all these operations. We observe that the element-wise multiplication, even though it produces accurate density maps, does not offer the best results. The summation is a linear operation that shows satisfying results. It also has the advantage that it does not add channels to the resulting features, making this type of operation suitable for situations in which memory consumption and processing capabilities are constrained. The best performing operation is the concatenation, hence, for the remainder of our experiments, we use it as the merging operation of the short-cut connections.

\begin{figure}[t!]
\centering
\includegraphics[width=0.78\textwidth]{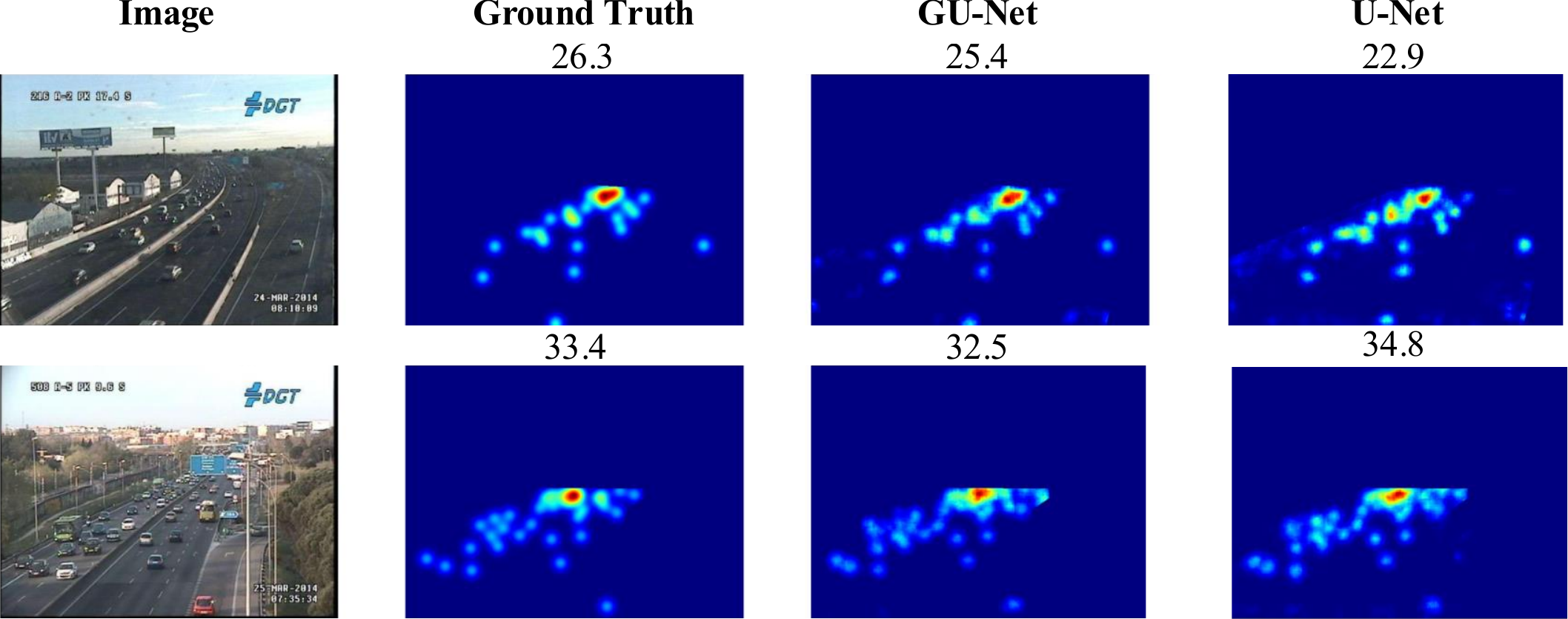}
\caption{\label{fig:qualitative-trancos}  \textsc{Trancos} qualitative results.}
\end{figure}

We now compare in Table~\ref{tab:trancos_art} our solutions with the state-of-the-art in the \textsc{Trancos} data set. Shallow methods are listed in the upper part of the table. In the second part we find the deep learning methods. The last two rows show the results obtained by the U-Net and the GU-Net models. First, our GU-Net achieves better results than the U-Net base architecture. The improvements are consistent across the different $\mathtt{GAME}$ settings used. This validates our hypothesis about the proposed learnable gating units. Second, in our analysis one can appreciate that the shallow methods are outperformed by deep learning based approaches. Among the deep models, the recent work of \cite{yuhong2018_csrnet} gets the best performance for $\mathtt{GAME}(0)$ (or $MAE$). However, it reports high errors for $\mathtt{GAME}(2)$ and $\mathtt{GAME}(3)$. This indicates that the model is probably under-counting and over-counting in different regions of the image, but the count over the whole image is compensated. Overall, the best vehicle counting performance for both $\mathtt{GAME}(2)$ and $\mathtt{GAME}(3)$ is obtained by our GU-Net. Figure \ref{fig:qualitative-trancos} shows a few qualitative results of the proposed models.

\begin{table*}[t!]
\small
\centering
\caption{Results comparison for \textsc{Trancos} dataset.}
\label{tab:trancos_art}
\begin{tabular}{|l|c|c|c|c|}
\hline
\multicolumn{5}{c}{\textsc{Trancos}}                                         \\
\hline
\hline
Model                           & GAME 0 & GAME 1 & GAME 2 & GAME 3 \\
\hline
\hline
Reg. Forest\cite{fiaschi2012}   & 17.77  & 20.14  & 23.65  & 25.99  \\
\hline
MESA \cite{lempitsky2010}       & 13.76  & 16.72  & 20.72  & 24.36  \\
\hline
\hline
Hydra 3s \cite{Onoro_2016_ECCV} & 10.99  & 13.75  & 16.69  & 19.32  \\
\hline
FCN-ST \cite{Zhang_2017_CVPR}   & 5.47   & -      & -      & -      \\
\hline
CSRNet \cite{yuhong2018_csrnet} & \textbf{3.56}   & \textbf{5.49}   & 8.57  & 15.04  \\
\hline
\hline
U-Net                           & 4.58   & 6.69   & 8.69   & 10.83  \\
\hline
GU-Net                          & 4.44   & 5.84   & \textbf{7.34}   & \textbf{9.29}   \\
\hline
\end{tabular}
\end{table*}

\begin{figure}[t!]
\centering
\subfigure[] {
  \includegraphics[width=0.45\linewidth]{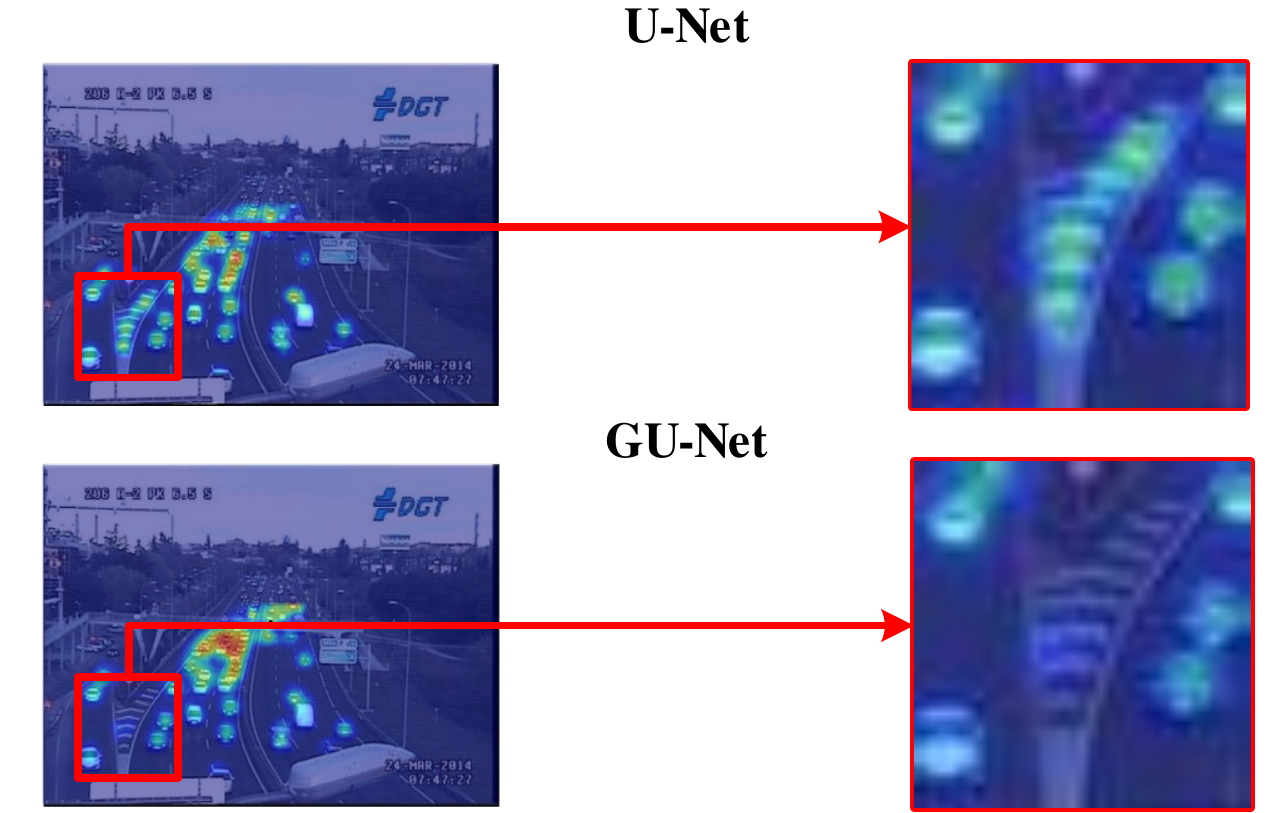}
  \label{fig:qualitative-errors}
}\hspace{5mm}
\subfigure[] {
  \includegraphics[width=0.45\linewidth]{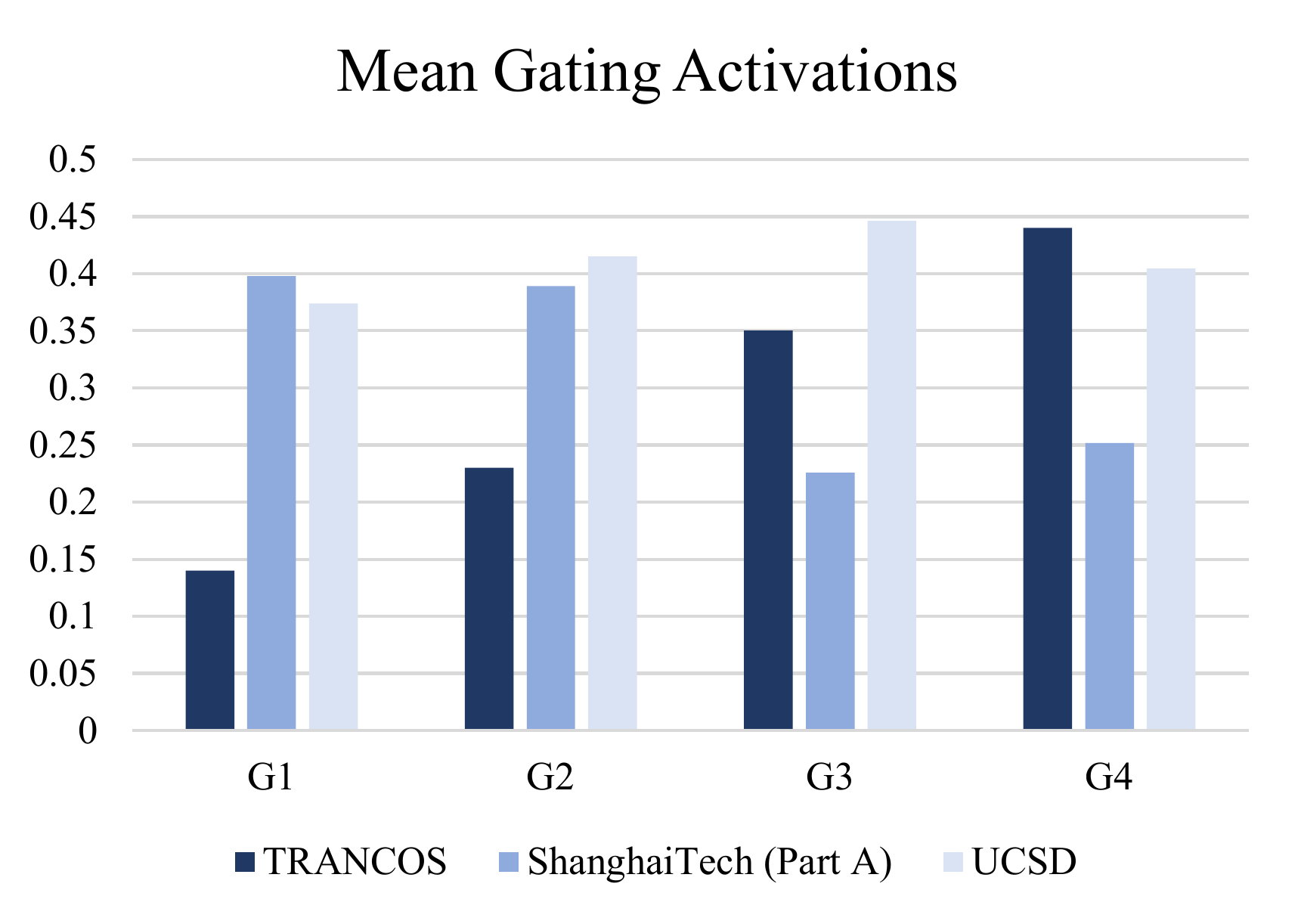}
  \label{fig:gating-scores}
}
\caption{a) Qualitative error analysis between U-Net and GU-Net. (b) Mean activation scores of the gating units of the analyzed datasets.}
\label{fig:general_diagram}
\end{figure}

Adding learnable short-cut units helps to determine the strength of the information that is passed to the later construction layers. It actually increases the robustness of the final model by blocking information that is not relevant for these layers. Intuitively, in a deep convolutional neural network, the first layers learn low-level features such as edges, textures, or colors. The later layers are able to capture more complex feature patterns such as eyes, legs, etc. Therefore, when low-level features are combined with high-level features, the resulting feature map might contain irrelevant information, potentially adding noise. As an example, in Figure~\ref{fig:qualitative-errors} we present a situation that clearly shows how a low level texture is adding noise. The U-Net is confusing the road lines as cars while the GU-Net can efficiently handle the problem. Our learnable short-cut units learn to detect these situation and effectively block the forwarded information through the short-cut connections. To explore this hypothesis more thoroughly, we measure the mean values of the activation scalars (the output of the sigmoids) of the four learnable short-cut connections in the GU-Net for several data sets. Figure \ref{fig:gating-scores} depicts the results of this experiment. It clearly shows that the effect of the short-cut units is data dependent, that is, the learnable short-cut units automatically adapt to the data set under consideration.

\subsection{Crowd counting}
In addition to the previous experiments, we also include a detailed comparison of the U-Net and GU-Net architectures using \textsc{ShanghaiTech}~\cite{zhang2016} and \textsc{UCSD} \cite{chan2008}, two publicly available data sets for the problem of crowd counting.

\textsc{ShanghaiTech}~\cite{zhang2016} contains 1.198 annotated images with a total of 330.165 persons. The dataset consists of two parts: Part A contains images randomly crawled on the Internet, and Part B is made of images taken from the metropolitan areas of Shanghai. Both sets are divided into training and testing subsets, where: i) Part A contains 300 training images and 182 test images; ii) Part B consists of 400 training images and 316 test images. In addition to the standard sets, we have created our own validation set for each part, by randomly taking 50 images out of the training sets. We resize the images to have a maximum height or width of 380 pixels, and we generate the ground truth density maps by placing a Gaussian kernel on each annotated position with a standard derivation of $\sigma = 4$. We follow the training procedure described in section~\ref{sec:setup}. 

Table~\ref{tab:shanghai_art} shows the results obtained for this dataset. First, one can again observe how the GU-Net improves the results of the standard U-Net baseline approach. In the Part A of the dataset, the CSRNet \cite{yuhong2018_csrnet} approach gets the best performance. It is followed by the CP-CNN~ \cite{Sindagi_2017_ICCV} model. However, in Part B, CSRNet reports the lowest error, but it is followed by our models. Note that both CP-CNN and CSRNet approaches are based on or include the VGG16 architecture. While VGG16 has around 16.8M of learnable parameters and is pretrained on Imagenet, our most complex model, \ie GU-Net, has only 2.8M parameters and is not pretrained on any external dataset.

\textsc{UCSD} is a standard benchmark in the crowd counting community. It provides 2000 video frames recorded by a camera capturing a single scene. The images have been annotated with a dot on each visible pedestrian. As in previous work~\cite{chan2008}, we train the models on frames 601 to 1400. The remaining frames are used for testing. For our experiments we sample $100$ frames uniformly at random from the set of training, and use them as our validation subset. To generate the ground truth density maps, we set the standard deviation of the normal distributions placed on the objects to $\sigma=5$. To train our models, we follow the procedure described in section~\ref{sec:setup}.

Table \ref{tab:ucsd_art} shows the results for the \textsc{UCSD} data set. Without hyper-parameter tuning, we are able to report state of the art results. Moreover, we observe how the GU-Net model consistently improves the U-Net. Our approaches are outperformed by CSRNet~\cite{yuhong2018_csrnet} and MCNN~\cite{zhang2016} models. However, note that our models are not pretrained, and that they are lighter in terms of parameters.

\begin{table}
\small
\centering
\caption{Crowd counting results for \textsc{ShanghaiTech} and UCSD datasets.}
\subfigure[\textsc{ShanghaiTech} state of the art.] {
\label{tab:shanghai_art}
\begin{tabular}{|l|c|c|c|c|}
\hline
\multicolumn{5}{c}{\textsc{ShanghaiTech}}                                  \\
\hline
\hline
\multirow{2}{*}{Model} & \multicolumn{2}{|c}{Part A} & \multicolumn{2}{|c|}{Part B}\\
\cline{2-5}
                                     & MAE & MSE & MAE & MSE \\
\hline
\hline
Cross-scene \cite{zhang_2015_CVPR}   & 181.8	& 277.7 & 32.0	& 49.8 \\
\hline
MCNN \cite{zhang2016}		          & 110.2	& 173.2 & 26.4	& 41.3 \\
\hline
\hline
CP-CNN \cite{Sindagi_2017_ICCV}      & 73.6	& \textbf{106.4} & 20.1	& 30.1 \\
\hline
CSRNet \cite{yuhong2018_csrnet}      & \textbf{68.2}	& 115.0 & \textbf{10.6}	& \textbf{16.0} \\
\hline
\hline
U-Net                                & 104.9	& 173.3 & 17.1	& 25.8 \\
\hline
GU-Net                               & 101.4	& 167.8 & 14.7 & 23.3 \\
\hline
\end{tabular}
} %
\subfigure[\textsc{UCSD} state of the art.] {
\label{tab:ucsd_art}
\begin{tabular}{|l|c|c|}
\hline
\multicolumn{3}{c}{\textsc{UCSD}} \\
\hline
\hline
Model                                            & MAE                     & MSE  \\
\hline
\hline
FCN-MT \cite{Zhang_2017_CVPR}                    & 1.67                    & 3.41 \\
\hline
Count-forest \cite{Pham_2015_ICCV}               & 1.61                    & 4.4  \\
\hline
Cross-scene \cite{zhang_2015_CVPR}               & 1.6                     & 5.97 \\
\hline
CSRNet \cite{yuhong2018_csrnet}                  & 1.16                    & 1.47 \\
\hline
MCNN \cite{zhang2016}                            & \textbf{1.07}                    & \textbf{1.35} \\
\hline
\hline
U-Net                                            & 1.28                    & 1.57 \\
\hline
GU-Net                                           & 1.25                    & 1.59 \\
\hline
\end{tabular}
}%
\vspace{-0.3cm}

\end{table}

%% file: conclusion.tex
\section{Conclusion}

In this paper, we have introduced the concept of learnable gating unit for convolutional-deconvolutional deep networks, like U-Net. Technically, our gating units work as short-cut connections, which can be directly learned from the training data, in order to optimize the flow of information between the layers they connect.

We demonstrate the effectiveness of this novel gating units for the problem of object counting. For doing so, we introduce a novel counting model named GU-Net, which is a U-Net architecture enriched with the learnable short-cut connections proposed. We have evaluated our solutions with three commonly used data sets for object counting. According to our results, GU-Net consistently outperforms the base U-Net architecture, and achieves a state-of-the-art performance.

\section*{Acknowledgments}
This work has received funding from the European Union's Horizon 2020 research and innovation programme under grant agreement No 761508, and the project PREPEATE, with reference TEC2016-80326-R, of the Spanish Ministry of Economy, Industry and Competitiveness. We acknowledge the support of NVIDIA Corporation with the donation of a GPU. Cloud computing resources were provided through a Microsoft Azure for Research Award.